\documentclass[letterpaper, 10 pt, conference]{ieeeconf}  

\IEEEoverridecommandlockouts                              

\usepackage{booktabs}
\usepackage{threeparttable}
\usepackage{multirow}
\usepackage{mathtools}
\usepackage{cite}
\usepackage{amsmath,amssymb,amsfonts}
\usepackage{algorithmic}
\usepackage{graphicx}
\usepackage{textcomp}
\usepackage{dsfont}
\usepackage[dvipsnames]{xcolor}
\graphicspath{{figures/}}
\usepackage{ragged2e}
\usepackage{tabularx}
\usepackage[ruled,linesnumbered]{algorithm2e}
\usepackage{subcaption}
\usepackage{bbding}
\usepackage{amssymb}
\usepackage{orcidlink}
\newcolumntype{R}[1]{>{\raggedleft\arraybackslash}p{#1}} 
\newcolumntype{L}[1]{>{\raggedright\arraybackslash}p{#1}} 
\newcolumntype{C}[1]{>{\centering\arraybackslash}p{#1}} 
\newcolumntype{Y}{>{\raggedright\arraybackslash}X}
\newcolumntype{Z}{>{\raggedleft\arraybackslash}X}
\newcolumntype{A}{>{\centering\arraybackslash}X}

\tikzstyle{startstop} = [rectangle, rounded corners, minimum width=3cm, minimum height=1cm, text centered, draw=black, fill=red!30]
\tikzstyle{io} = [trapezium, trapezium left angle=70, trapezium right angle=110, minimum width=3cm, minimum height=1cm, text centered, draw=black, fill=blue!30]
\tikzstyle{process} = [rectangle, minimum width=3cm, minimum height=1cm, text centered, draw=black, fill=orange!30]
\tikzstyle{arrow} = [thick,->,>=stealth]
\begin{document}
\pagestyle{empty}
\title{BiTrack: Bidirectional Offline 3D Multi-Object Tracking Using Camera-LiDAR Data}
\author{%
    Kemiao Huang~\orcidlink{0000-0003-2899-4222}, 
    Yinqi Chen~\orcidlink{0009-0000-4527-5751},
    Meiying Zhang~\orcidlink{0000-0002-3721-4315},
    Qi Hao~\orcidlink{0000-0002-2792-5965}%
    \thanks{Kemiao Huang and Yinqi Chen are co-first authors; Corresponding author: Meiying Zhang (e-mail: zhangmy@sustech.edu.cn), and Qi Hao (e-mail: hao.q@sustech.edu.cn)}%
    \thanks{All authors are within the Research Institute of Trustworthy Autonomous Systems, and Department of Computer Science and Engineering, Southern University of Science and Technology (SUSTech), Shenzhen 518055, China.}%
    \thanks{This work is jointly supported by the National Natural Science Foundation of China (62261160654), the Shenzhen Fundamental Research Program (JCYJ20220818103006012, KJZD20231023092600001), and the Shenzhen Key Laboratory of Robotics and Computer Vision (ZDSYS20220330160557001).}
}



\maketitle
\thispagestyle{empty}
\begin{abstract}
Compared with real-time multi-object tracking (MOT), offline multi-object tracking (OMOT) has the advantages to perform 2D-3D detection fusion, erroneous link correction, and full track optimization but has to deal with the challenges from bounding box misalignment and track evaluation, editing, and refinement. 
This paper proposes ``BiTrack'', a 3D OMOT framework that includes modules of 2D-3D detection fusion, initial trajectory generation, and bidirectional trajectory re-optimization to achieve optimal tracking results from camera-LiDAR data.
The novelty of this paper includes threefold: (1) development of a point-level object registration technique that employs a density-based similarity metric to achieve accurate fusion of 2D-3D detection results; (2) development of a set of data association and track management skills that utilizes a vertex-based similarity metric as well as false alarm rejection and track recovery mechanisms to generate reliable bidirectional object trajectories; (3) development of a trajectory re-optimization scheme that re-organizes track fragments of different fidelities in a greedy fashion, as well as refines each trajectory with completion and smoothing techniques. The experiment results on the KITTI dataset demonstrate that BiTrack achieves the state-of-the-art performance for 3D OMOT tasks in terms of accuracy and efficiency. 
\end{abstract}
\section{Introduction}

\begin{figure}
\centering
\includegraphics[width=\linewidth]{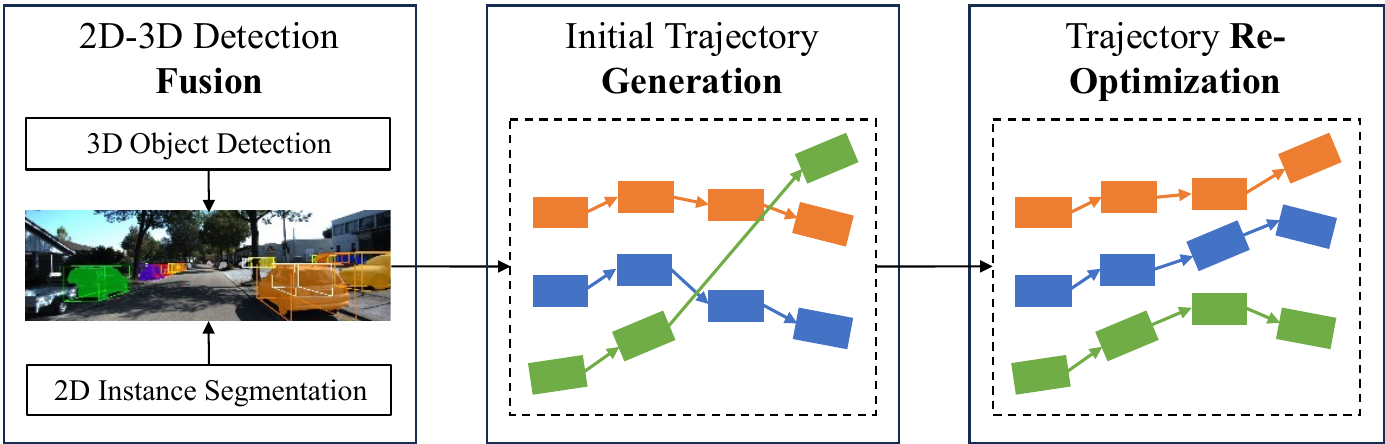}
\caption{A post-processing based offline 3D multi-object tracking framework with three main stages.}
\label{fig:intro}
\vspace{-0.7cm} 
\end{figure}

Many applications, such as movement analysis and dataset annotation, require high-accuracy object trajectories from offline multi-object tracking (OMOT). Real-time multi-object tracking (MOT) typically uses tracking-by-detection \cite{bewley2016simple,Weng2020_AB3DMOT,wu20213d,du2023strongsort} or joint detection and tracking \cite{huang2021joint}. OMOT generally prefers the former, as many post-processing and global optimization techniques rely on detection results. OMOT association frameworks fall into two categories: (1) optimization and clustering of detection results \cite{networkflows,mussp,AHC_ETE}, and (2) editing and refining initial trajectories \cite{ReMOT,TMOH,Qi_2021_CVPR,wu20213d,du2023strongsort}. Both depend on detection quality. The first category struggles with context inconsistency and computational instability due to the lack of sequential information, while the second guarantees performance by using sequential tracking results. Many methods improve 3D detection using 2D results, but are limited by the cascade pipeline \cite{shenoi2020jrmot} or 2D-3D registration errors \cite{pang2020clocs}. Therefore, to develop a post-processing-based OMOT framework (Fig. \ref{fig:intro}), the following technical challenges must be well dealt with:
\begin{enumerate}
\item \textbf{2D-3D object registration}. Sensor miscalibration, occlusion, detection errors, and bounding box misalignment introduce noise in post-fusion registration.
\item \textbf{Initial trajectory generation}. Association accuracy relies on object similarity and track management, but is hindered by complex motion, false alarms, and reappearances.
\item \textbf{Trajectory post-processing}. Involves (1) reorganizing trajectories via evaluation and association, and (2) refining individual trajectories through completion and regression, facing temporal contradictions, uncertainties, and high computational cost.
\end{enumerate}


\begin{table*}\scriptsize
    \caption{A methodological comparison between BiTrack and typical 2D-3D object detection and tracking methods}
    \label{tab:related-work}
    \centering
    \resizebox{0.9\textwidth}{!}{
    \begin{threeparttable}[c]
    \begin{tabularx}{\linewidth}{m{1.7cm}|m{1.4cm}|m{1.9cm}|m{3.5cm}|m{2.0cm}|m{3.0cm}|m{1.8cm}}
        \toprule
         & \multirow{3.5}{*}{Category} & \multirow{3.5}{*}{\shortstack{2D-3D Object\\ Fusion}} & \multicolumn{2}{c |}{Initial Trajectory Generation} & \multicolumn{2}{c}{Trajectory Post-Processing} \\
        \cmidrule(lr){4-5}\cmidrule(lr){6-7}
        & & & Similarity Evaluation & Track Management & Multi-Trajectory Fusion & Single-Trajectory Refinement \\ 
        \midrule
        CLOCs\cite{pang2020clocs} & 3D Detection & Box-Level Fusion & - & - & - & - \\
        \hline
        AB3DMOT\cite{Weng2020_AB3DMOT} & 3D MOT & $\times$ & 3D IoU/Center Distance & Online Hit-Miss & $\times$ & $\times$ \\
        \hline
        JRMOT\cite{shenoi2020jrmot} & 3D MOT & Frustum-Based Detection Pipeline & Appearance / 2D IoU / 3D IoU & Online Hit-Miss & $\times$ & $\times$ \\
        \hline
        AHC\_ETE\cite{AHC_ETE} & 2D OMOT & $\times$ & Appearance / 2D Center Distance & $\times$ & Hierarchical Clustering & $\times$ \\
        \hline
        TMOH\cite{TMOH} & 2D OMOT & $\times$ & Appearance & Occlusion-Aware Online Miss & Bidirectional Trajectory Replacement & $\times$ \\
        \hline
        ReMOT\cite{ReMOT} & 2D OMOT & $\times$ & Appearance / 2D Center Distance & Online Hit-Miss & Forward Tracklet Split-Merge & $\times$\\
        \hline
        StrongSORT\cite{du2023strongsort} & 2D OMOT & $\times$ & Appearance \& 2D Center Distance & Online Hit-Miss & Tracklet Association & LI \& GPR \\
        \hline
        3DAL\cite{Qi_2021_CVPR} & 3D OMOT & $\times$ & 3D IoU/Center Distance & Online Hit-Miss & $\times$ & Deep Learning Regression \\
        \hline
        PC3T\cite{wu20213d} & 3D OMOT & $\times$ & 3D Center Distance & Online Miss & $\times$ & Kalman Filter Prediction \\
        \hline
        BiTrack (ours) & 3D OMOT & Point-Level Fusion & 3D Normalized Center Distance & Improved Offline Hit-Miss & Bidirectional Trajectory Split-Merge & LI \& CSA \& GPR \\
        \bottomrule
    \end{tabularx}
    \begin{tablenotes}
        \item ``$\times$'': unsupported functionality, ``LI'': linear interpolation, ``CSA'': confidence-based size averaging, ``GPR'': Gaussian process regression
    \end{tablenotes}
    \end{threeparttable}
    }
    \vspace{-0.5cm} 
\end{table*}

Several detection fusion methods \cite{pang2020clocs} use 2D IoUs between 2D and 3D bounding boxes as fusion cues, but this can cause ambiguities due to occlusions. Most real-time MOT methods \cite{bewley2016simple,Weng2020_AB3DMOT,shenoi2020jrmot,du2023strongsort} use bounding box-based metrics for object similarity and life cycle mechanisms for track management, yet few address metric limitations or detection errors. Many OMOT methods refine trajectories with clustering \cite{AHC_ETE}, association \cite{du2023strongsort} or regression \cite{Qi_2021_CVPR,du2023strongsort}, but struggle to re-organize trajectories using global sequential information.

This paper presents ``BiTrack'', an OMOT framework which enables robust 2D-3D detection fusion, reliable initial trajectory generation, and efficient trajectory re-optimization. The contributions of this work include:
\begin{enumerate}
    \item Development of a 2D-3D detection fusion module that utilizes point-level representations and object point densities to achieve robust object registration.
    \item Development of a trajectory generation module that employs a scale-balanced object similarity metric and an offline-oriented track management mechanism to achieve reliable 3D MOT.
    \item Development of a trajectory re-optimization module that utilizes a priority-based fragment optimization, trajectory completion and smoothing skills for efficient bidirectional fusion and trajectory refinement.
    \item Release of the source code of this work at \url{https://github.com/Kemo-Huang/BiTrack}.
\end{enumerate}


\begin{figure*}[t]
\centering
\includegraphics[width=0.9\linewidth]{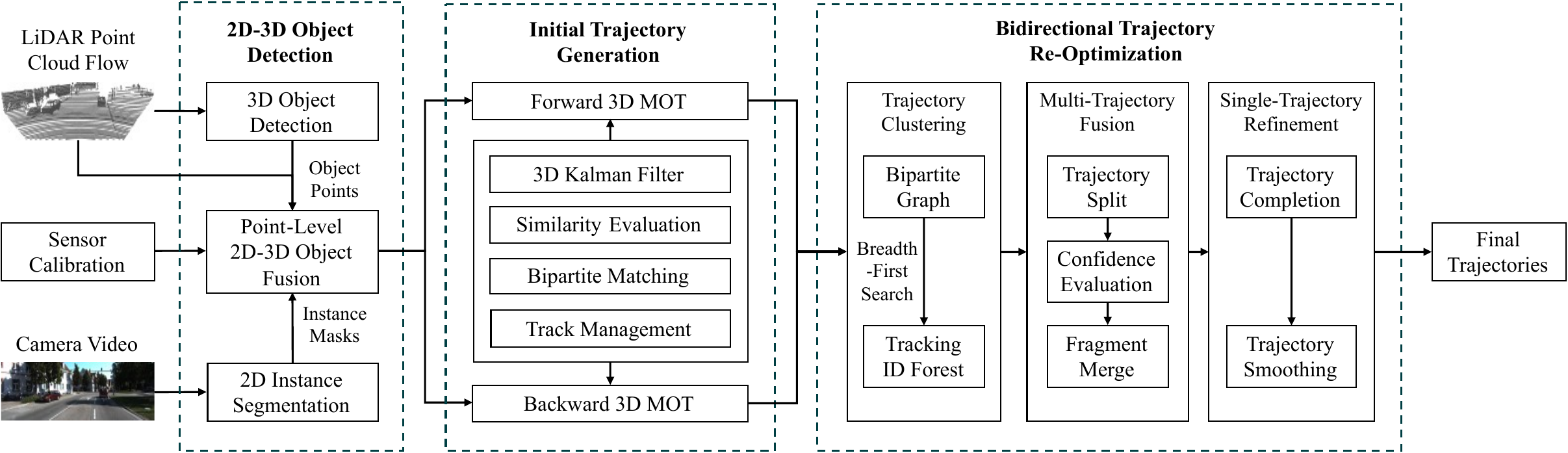}
\caption{The system architecture of BiTrack.}
\label{fig:system}
\vspace{-0.5cm}
\end{figure*}

\section{Related Work}
\label{sec:related-work}
Typical methods for 2D-3D object detection, real-time MOT and OMOT are summarized in TABLE \ref{tab:related-work}. There are many multi-modal methods in 3D object detection\cite{Qi_2018_CVPR,pang2020clocs} and 3D MOT\cite{shenoi2020jrmot}. JRMOT\cite{shenoi2020jrmot} adopts the frustum-based fusion method\cite{Qi_2018_CVPR} for cascade 2D-3D detection, where the 2D detection becomes the bottleneck. CLOCs\cite{pang2020clocs} uses separate detection branches and re-evaluates 3D detection confidences using 2D results. However, the 2D-3D object correspondences are based on 2D IoU, which may suffer from object occlusions. 

IoU and the center distance are two prevalent object motion similarity metrics for 2D and 3D MOT\cite{Weng2020_AB3DMOT,shenoi2020jrmot,AHC_ETE,Qi_2021_CVPR,ReMOT,wu20213d,du2023strongsort}. Both metrics provide incomplete spatial information, while 2D appearance similarities are hard to be combined with the latter due to the scale gap\cite{du2023strongsort}. In addition, hit-miss schemes are commonly used in the track management\cite{Weng2020_AB3DMOT,shenoi2020jrmot,Qi_2021_CVPR,ReMOT,wu20213d,du2023strongsort} but most are only designed for online applications. 

Many OMOT methods\cite{TMOH,ReMOT,du2023strongsort,Qi_2021_CVPR,wu20213d} perform trajectory post-processing based on real-time MOT, requiring sequential association. Methods like min-cost flow\cite{networkflows} and hierarchical clustering\cite{AHC_ETE} merge detections globally without sequential cues, but suffer from high computation and inconsistent object similarities. ReMOT\cite{ReMOT} uses a sliding window to re-evaluate 2D object similarities for tracklet split-merge, while TMOH\cite{TMOH} assembles forward and backward sequences, addressing object link contradictions by replacing the entire trajectory. Single-trajectory refinement can be done using physical\cite{du2023strongsort,wu20213d} or deep learning models\cite{Qi_2021_CVPR}.

BiTrack follows the refinement scheme for both detection and tracking. In contrast to previous works, this work provides the features of point-level detection fusion, robust motion-based initial trajectory generation, as well as split-merge based bidirectional trajectory re-optimization. 

\section{System Setup and Problem Statement}
\label{sec:system}

BiTrack can be divided into three main modules: (1) 2D-3D object detection, (2) initial trajectory generation, and (3) bidirectional trajectory re-optimization, as shown in Fig. \ref{fig:system}. The whole pipeline is fully automatic. 


For data pre-processing, object detectors localize 2D objects $\boldsymbol{\mathcal{D}}^\text{2D}$ and 3D objects $\boldsymbol{\mathcal{D}}^\text{3D}$ with detection confidences $\boldsymbol{\mathcal{C}}^\text{2D}$ and $\boldsymbol{\mathcal{C}}^\text{3D}$ in separate branches. BiTrack performs OMOT as follows. 

\textbf{Firstly}, the 2D-3D object fusion module performs the object registration between $\boldsymbol{\mathcal{D}}^\text{2D}$ and $\boldsymbol{\mathcal{D}}^\text{3D}$ using camera intrinsic parameters $\boldsymbol{K}$, camera-LiDAR extrinsic parameters $[\boldsymbol{R} \mid \boldsymbol{t}]$, and 2D-3D detection similarities $\boldsymbol{\mathcal{W}}^\text{D} \in \mathbb{R}^+$. The 2D-3D object registration is addressed as a complete bipartite graph matching problem to solve the assignment matrix $\boldsymbol{\mathcal{X}}^\text{D}$:
\begin{equation}\scriptsize
\label{eq:2d3dfusion}
\begin{gathered}
    \boldsymbol{\mathcal{X}}^{\text{D}*} = \mathop{\arg\max}_{X^\text{D}}
    \sum_{D^\text{2D}_i \in \boldsymbol{\mathcal{D}}^\text{2D}} \sum_{D_j^\text{3D} \in \boldsymbol{\mathcal{D}}^\text{3D}}
    W^\text{D}_{i,j} X^\text{D}_{i,j} \\
    \text{s.t. } \begin{dcases}
        \sum_{D^\text{3D}_j\in \boldsymbol{\mathcal{D}}^\text{3D}}X^\text{D}_{i,j} \leq 1 & \forall D^\text{2D}_i \in \boldsymbol{\mathcal{D}}^\text{2D} \\
        \sum_{D^\text{2D}_i\in \boldsymbol{\mathcal{D}}^\text{2D}}X^\text{D}_{i,j} \leq 1 & \forall D^\text{3D}_j \in \boldsymbol{\mathcal{D}}^\text{3D} \\
        X^\text{D}_{i,j} \in \{0, 1\} & \forall D^\text{2D}_i \in \boldsymbol{\mathcal{D}}^\text{2D}, \forall D^\text{3D}_j \in \boldsymbol{\mathcal{D}}^\text{3D} \\
    \end{dcases}
\end{gathered}
\end{equation}
The final 3D detection inputs for tracking $\boldsymbol{\mathcal{D}} = \{D^\text{3D}_i \in \boldsymbol{\mathcal{D}}^\text{3D} \mid f^\text{D}(D^\text{3D}_i) = 1 \}$ are selected from $\boldsymbol{\mathcal{D}}^\text{3D}$ according to $\boldsymbol{\mathcal{C}}^\text{3D}$, $ \boldsymbol{\mathcal{W}}^\text{D}$, and $ \boldsymbol{\mathcal{X}}^{\text{D}*}$ with the detection decision function $f^\text{D}(\cdot)$:
\scriptsize
\begin{multline}
    f^\text{D}(D^\text{3D}_i) = \mathds{1}\bigg[\left(C^\text{3D}_i \geq \theta^\text{det}\right)
    \lor \left(\exists D^\text{2D}_j \in \boldsymbol{\mathcal{D}}^\text{2D}:  W^\text{D}_{i,j} X_{i,j}^{\text{D}*} \geq \alpha\right) \bigg]
\end{multline}
\normalsize
where $\theta^\text{det}$ is the detection confidence threshold, $\alpha$ is the similarity threshold, and $\mathds{1}[\cdot] \rightarrow \{0, 1\}$ is the indicator function.  

\textbf{Secondly}, the initial trajectory generation module performs the association between detected objects $\boldsymbol{\mathcal{D}}$ and predicted objects $\boldsymbol{\mathcal{\hat{D}}}$ with detection-prediction similarities $\boldsymbol{\mathcal{W}}^\text{T} \in \mathbb{R}^+$. The association is also formulated as the bipartite matching problem to solve the assignment matrix $\boldsymbol{\mathcal{X}}^{\text{T}}$:
\begin{equation}\scriptsize
\label{eq:mot}
\begin{gathered}
    \boldsymbol{\mathcal{X}}^{\text{T}*} = \mathop{\arg\max}_{X^\text{T}}
    \sum_{D_i \in \boldsymbol{\mathcal{D}}} \sum_{\hat{D}_j \in \boldsymbol{\mathcal{\hat{D}}}}
    W^\text{T}_{i,j} X^\text{T}_{i,j} \\
    \text{s.t. } \begin{dcases}
        \sum_{\hat{D}_j\in \boldsymbol{\mathcal{\hat{D}}}}X^\text{T}_{i,j} \leq 1 & \forall D_i \in \boldsymbol{\mathcal{D}} \\
        \sum_{D_i\in \boldsymbol{\mathcal{D}}} X^\text{T}_{i,j} \leq 1 & \forall \hat{D}_j \in \boldsymbol{\mathcal{\hat{D}}} \\
        X^\text{T}_{i,j} \in \{0, 1\} & \forall D_i \in \boldsymbol{\mathcal{D}}, \forall \hat{D}_j \in \boldsymbol{\mathcal{\hat{D}}} \\
    \end{dcases}
\end{gathered}
\end{equation}
Tracked objects $\boldsymbol{\mathcal{D}}' = \{D_i \in \boldsymbol{\mathcal{D}} \mid g(D_i) = 1\}$ are sequentially selected from $\boldsymbol{\mathcal{D}}$ according to $\boldsymbol{\mathcal{W}}^\text{T}$ and $\boldsymbol{\mathcal{X}}^{\text{T}*}$ with the track decision function $f^\text{T}(\cdot)$ and the track management function $g(\cdot)$ in case of false alarms and object re-appearances:
\begin{equation}\scriptsize
    f^\text{T}(D_i) = \mathds{1}\left[\exists \hat{D}_j \in \boldsymbol{\mathcal{\hat{D}}}: W^\text{T}_{i,j} X_{i,j}^{\text{T}*} \geq \beta \right ]
\end{equation}
\scriptsize
\begin{align}
    \forall X_{i,j}^{\text{T}*} = 1: n^\text{hit}(D_i) & = \begin{cases}
        n^\text{hit}(D_i) & \text{if } f^\text{T}(D_i) = 0 \\
        n^\text{hit}(D_j) + 1 & \text{if } f^\text{T}(D_i) = 1
    \end{cases} \\
    n^\text{miss}(D_i) & = \begin{cases}
        n^\text{miss}(D_i) + 1 & \text{if } f^\text{T}(D_i) = 0 \\
        0 & \text{if } f^\text{T}(D_i) = 1
    \end{cases}
\end{align}
\normalsize
\begin{equation}\scriptsize
    g(D_i) = \begin{cases}
    1 & \text{if } \left( n^\text{hit}(D_i) \geq \theta^\text{hit} \right) \land \left( n^\text{miss}(D_i) = 0 \right) \\
    -1 & \text{if } n^\text{miss}(D_i) \geq \theta^\text{miss} \\
    0 & \text{otherwise}
    \end{cases}
\end{equation}
where $\beta$ is the tracking similarity threshold, $n^\text{hit}, n^\text{miss} \in \mathbb{N}_0$ are the times of hit and miss, and $\theta^\text{hit}, \theta^\text{miss} \in \mathbb{N}_0$ are thresholds. Tracking IDs can only be assigned to confirmed objects $\left(g(\cdot) = 1\right)$. Object candidates $\left(g(\cdot) = 0\right)$ are used for temporary object prediction while dead objects $\left(g(\cdot) = -1\right)$ are permanently removed. The confirmed objects with the same IDs are connected with object links $\boldsymbol{\mathcal{L}}$ to generate trajectories $\boldsymbol{\mathcal{T}}$. The same detection input $\boldsymbol{\mathcal{D}}$ is used to perform MOT in both forward and backward time directions to generate trajectories $\boldsymbol{\mathcal{T}}^\text{A} = \left \{ \boldsymbol{\mathcal{D}}^\text{A}, \boldsymbol{\mathcal{L}}^\text{A} \right \}$ and $\boldsymbol{\mathcal{T}}^\text{B} = \left \{ \boldsymbol{\mathcal{D}}^\text{B}, \boldsymbol{\mathcal{L}}^\text{B} \right \}$. 

\begin{figure}
\centering
\includegraphics[width=1.0\linewidth]{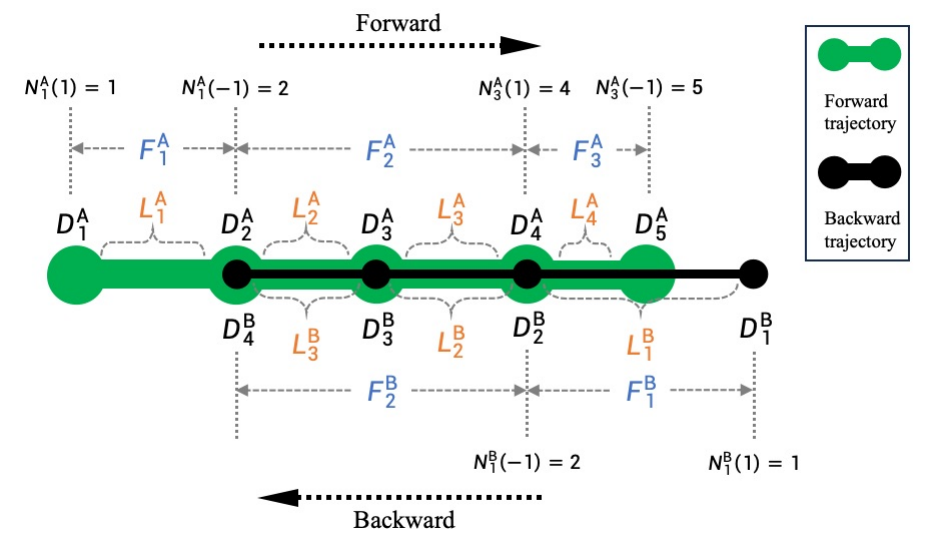}
\caption{A visual explanation of the symbol definitions for bidirectional trajectory fusion. Best viewed in color.}
\label{fig:symbol}
\vspace{-0.5cm} 
\end{figure}

\textbf{Finally}, the trajectory re-optimization module performs the trajectory fusion and refinement. Fig. \ref{fig:symbol} demonstrates the symbol definitions. $\boldsymbol{\mathcal{T}}^\text{A}$ and $\boldsymbol{\mathcal{T}}^\text{B}$ are split into track fragments $\boldsymbol{\mathcal{F}}^\text{A}$ and $\boldsymbol{\mathcal{F}}^\text{B}$ by the common links $\boldsymbol{\mathcal{L}}^\text{A} \cap \boldsymbol{\mathcal{L}}^\text{B}$. The fragments are selected via the optimization of fragment priorities under the condition of exclusive object frames:
\begin{equation}\scriptsize
\label{eq:bi-opt}
\begin{gathered}
    \boldsymbol{\mathcal{F}}^* = \mathop{\arg\max}_F \sum_{F_i \in \boldsymbol{\mathcal{F}}^\text{A} \cup \boldsymbol{\mathcal{F}}^\text{B} } h(F_i)\\
    \text{s.t. } \forall T_i \in \boldsymbol{\mathcal{T}}, \forall D_j \neq D_k \in T_i: t(D_j) \neq t(D_k)  
\end{gathered}
\end{equation}
where $t(\cdot)$ is the frame of an object and $h(\cdot)$ is the fragment priority function:
\begin{equation}\scriptsize
    h(F_i) = \begin{cases}
        Z & \text{if } F_i \in \boldsymbol{\mathcal{F}}^\text{A} \cap \boldsymbol{\mathcal{F}}^\text{B} \\
        \max\left \{N^\text{A}_i(1), N^\text{A}_i(-1) \right \} & \text{if } F_i \in \boldsymbol{\mathcal{F}}^\text{A} - \boldsymbol{\mathcal{F}}^\text{B} \\
        \max\left \{N^\text{B}_i(1), N^\text{B}_i(-1) \right \} & \text{if } F_i \in \boldsymbol{\mathcal{F}}^\text{B} - \boldsymbol{\mathcal{F}}^\text{A}
    \end{cases}
\end{equation}
where $Z \gg N$ is a large integer, $N_i(1)$ and $N_i(-1)$ stand for the sequential index of the first and last object of $F_i$ in the entire trajectory respectively. The fused trajectories $\boldsymbol{\mathcal{T}}^\text{C} = \{ \boldsymbol{\mathcal{D}}^\text{C}, \boldsymbol{\mathcal{L}}^\text{C} \}$ are obtained from the optimization results $\boldsymbol{\mathcal{F}}^*$. The object localization results can be further regressed using the temporal information:
\begin{equation}\scriptsize
\label{eq:refine}
    \boldsymbol{\mathcal{D}}^\text{C} \xrightarrow{\boldsymbol{\mathcal{L}}^\text{C}}\boldsymbol{\mathcal{D}}^{+}
\end{equation}
The final trajectory outputs become $\boldsymbol{\mathcal{T}}^+ = \{\boldsymbol{\mathcal{D}}^+, \boldsymbol{\mathcal{L}}^\text{C} \}$.

Therefore, this work is to solve the following problems: 
\begin{enumerate}
    \item How to evaluate 2D-3D detection similarities $\boldsymbol{\mathcal{W}}^\text{D}$?
    \item How to evaluate detection-prediction similarities $\boldsymbol{\mathcal{W}}^\text{T}$ and design the hit-miss thresholds $\{\theta^\text{hit}, \theta^\text{miss}\}$?
    \item How to perform the optimization of eq. \eqref{eq:bi-opt} and the refinement of eq. \eqref{eq:refine}?
\end{enumerate}

\section{Proposed Methods}
\label{sec:proposed-methods}
\subsection{2D-3D Object Registration via Points}
This work suggests that 2D IoU has three shortages for 2D-3D object fusion: (1) the PVs of 3D boxes are larger than 2D boxes for occluded objects, (2) the PVs of 3D boxes could overlap, and (3) boxes contain space at corners. However, the point density metric between 2D segmentation masks and 3D object points can perfectly tackle those problems.

\subsubsection{Point-Level Segmentation}
To obtain pixel-level object masks, 2D instance segmentation is required. On the other hand, the 3D bounding boxes obtained from 3D object detection can be directly used to crop LiDAR points for efficiency because 3D objects rarely overlap, especially for rigid bodies. Although cropping LiDAR points within bounding boxes may cost more computations, the operations of multiple data frames can be paralleled by multi-processing for faster execution thanks to the offline setting.

\subsubsection{Point-Level Association}
After removing background, the correspondence between 2D and 3D objects should be robust against object occlusions. However, sensor miscalibration and detection inaccuracy may still produce registration noises, e.g., a 2D instance mask can include the projected points of multiple 3D objects. Thus, the 2D-3D object similarity should be quantified and the optimization should be performed. This work uses the number of overlapped pixels as $\boldsymbol{\mathcal{W}}^D$ and the Hungarian algorithm\cite{kuhn1955hungarian} to solve the eq. \eqref{eq:2d3dfusion}.

\begin{figure}
\centering
\begin{subfigure}{0.49\linewidth}
    \centering
    \includegraphics[width=0.8\linewidth]{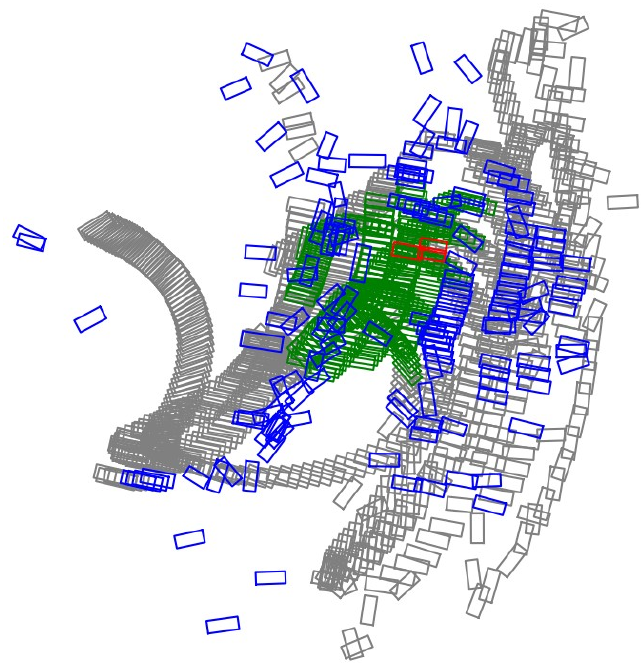}
    \caption{Original detection results\cite{Wu_2023_CVPR}}
\end{subfigure}
\hfill
\begin{subfigure}{0.49\linewidth}
    \centering
    \includegraphics[width=0.8\linewidth]{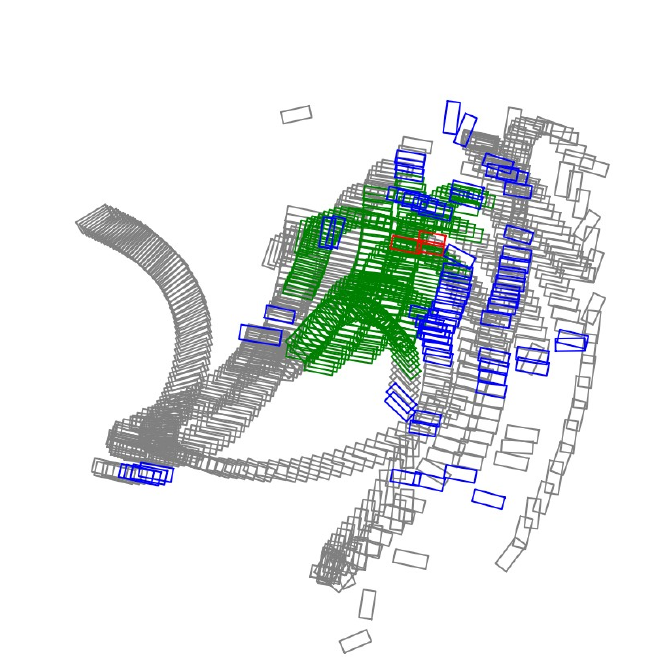}
    \caption{After 2D-3D object fusion}
\end{subfigure}
\caption{The bird's-eye view of 2D-3D object fusion results on a sequence. The \textcolor{gray}{\textbf{gray}} boxes are the objects that the dataset doesn't care. The \textcolor{Green}{\textbf{green}} boxes are true positives. The \textcolor{blue}{\textbf{blue}} boxes are false positives. The \textcolor{red}{\textbf{red}} boxes are false negatives.}
\label{fig:2d3dfusion}
\vspace{-0.5cm} 
\end{figure}

Fig. \ref{fig:2d3dfusion} demonstrates the effect of 2D-3D object fusion. The original detection results are generated by the SOTA method VirConv\cite{Wu_2023_CVPR} on the KITTI\cite{Geiger2012CVPR} dataset. The proposed 2D-3D fusion method can effectively reduce false alarms, while high-quality detection results can be successfully selected for the trajectory generation stage. The whole fusion procedure is shown in Algorithm \ref{algo:det-fusion}.

\subsection{Reliable Initial Trajectory Generation}
Most 3D MOT methods follow the SORT\cite{bewley2016simple} framework. This work improves the simple 3D MOT baseline\cite{weng20203d} with the following techniques for reliable trajectory generation.

\subsubsection{Integrated Object Motion similarity Metric}
\label{subsubsec:sim-metric}
Both 3D IoU and the center distance (CD) metrics have shortages: (1) 3D IoU is too sensitive to object rotation errors, (2) CD does not utilize object size and rotation cues, (3) 3D IoU is incomparable (zero) for separate objects, and (4) the numerical value of CD has no upper bound. To complement these two metrics, this work proposes the normalized center distance (NCD) as the geometric cost. Specifically, the proposed NCD similarity metric $\boldsymbol{\mathcal{W}}^\text{T}$ for $D_i$ and $\hat{D}_j$ is defined as:
\begin{equation}\scriptsize
\label{eq:norm-center-dis}
    W^\text{T}_{i,j} = 1 - \frac{d(c_i, \hat{c}_j)}{\max \left \{d(v_{i,k}, \hat{v}_{j,l}) \mid v_{i,k} \in \boldsymbol{\mathcal{V}}_i, \hat{v}_{j,l} \in \boldsymbol{\mathcal{\hat{V}}}_j \right \}}
\end{equation}
where $d( \cdot )$ denotes the Euclidean distance, $c$ denotes the center of the bounding box, $\boldsymbol{\mathcal{V}}$ denotes bounding box vertices, $\hat{c}_j$ and $\boldsymbol{\mathcal{\hat{V}}}_j$ are predicted values from the Kalman filter. The NCD metric provides two main advantages for object similarity evaluation: (1) simultaneous utilization of bounding box positions, sizes, and rotations; (2) normalized numerical value for easy combination with other similarities (e.g., the weighted sum with the cosine similarity between appearance embeddings).

\subsubsection{Previous Tracklet Recovery and Double Miss Thresholds}
Online tracking results should be reported sequentially but OMOT can break this restriction. Once objects are confirmed, their previous tracklets can be recovered to generate more complete trajectories. Additionally, based on the intuition that false alarms usually do not appear continually, an extra miss threshold is set for track candidates so that false alarms can be quickly rejected. 



\subsubsection{Velocity Re-initialization}
Most SORT-based methods\cite{du2023strongsort,weng20203d,wu20213d} simply initialize object linear velocities as zeros and leave the estimation to the Kalman filter in later updates. However, this inaccurate prior knowledge makes the Kalman filter hard to converge and leads to inaccurate predictions in the early stage. This work proposes to re-initialize the states of objects after they are firstly matched with new measurements. Specifically, the static state is updated exactly as the new measurement and the linear velocity is re-initialized as the translation divided by the passed frames. All state covariances are updated as usual.
\begin{algorithm}[!t]
\scriptsize
    \caption{Point-level fusion between 2D instance masks and 3D bounding boxes}
    \label{algo:det-fusion}
    \KwIn{\parbox[t]{\linewidth}{
        2D instance masks $\boldsymbol{\mathcal{D}}^\text{2D}$;\\
        3D bounding boxes $\boldsymbol{\mathcal{D}}^\text{3D}$;\\
        3D detection confidences $\boldsymbol{\mathcal{C}}^\text{3D}$;\\
        LiDAR point clouds $\boldsymbol{\mathcal{P}}^{\text{LiDAR}}$;\\
        Sensor calibration parameters $\left\{\boldsymbol{K},  [\boldsymbol{R} \mid \boldsymbol{t}]\right\}$;
        \vspace{1mm}
    }}
    \KwOut{the selected 3D bounding boxes $\boldsymbol{\mathcal{D}}$;}
    $\boldsymbol{\mathcal{P}}^\text{3D} \gets $ Crop the LiDAR points within $\boldsymbol{\mathcal{D}}^\text{3D}$ from $\boldsymbol{\mathcal{P}}^{\text{LiDAR}}$\;
    $\boldsymbol{\mathcal{P}}^\text{2D} \gets $ Project $\boldsymbol{\mathcal{P}}^\text{3D}$ to camera images using $\left\{\boldsymbol{K},  [\boldsymbol{R} \mid \boldsymbol{t}]\right\}$\;
    $\boldsymbol{\mathcal{P}}^\text{2D} \gets $ Align $\boldsymbol{\mathcal{P}}^\text{2D}$ to image pixels\;
    $\boldsymbol{\mathcal{W}}^\text{D} \gets \left \{\left|P^\text{2D}_i \cap D^\text{2D}_j\right| : P^\text{2D}_i \in \boldsymbol{\mathcal{P}}^\text{2D}, D^\text{2D}_j \in \boldsymbol{\mathcal{D}}^\text{2D} \right \}$\;
    $\boldsymbol{\mathcal{X}}^{\text{D}*} \gets $ Solve the eq. \eqref{eq:2d3dfusion} using the Hungarian algorithm\;
    $\boldsymbol{\mathcal{D}} \gets \left\{D^\text{3D}_i \mid f^\text{D}(D^\text{3D}_i) = 1, D^\text{3D}_i \in \boldsymbol{\mathcal{D}}^\text{3D} \right \} $\;
    \Return $\boldsymbol{\mathcal{D}}$
\end{algorithm}
\vspace{-0.3cm}
\subsection{Efficient Trajectory Re-Optimization}
Forward and backward trajectory integration realizes false alarms and ID switches, while sequential bounding box refinement realizes false negatives and regression errors. This work proposes three efficient post-processing steps for MOT to improve the link accuracy, the bounding box integrity, and the trajectory smoothness.

\subsubsection{Bidirectional Trajectory Clustering}
Clustering forward and backward trajectories into groups for fusion is based on equal bounding boxes. By ordering bounding boxes sequentially, the “two pointers” technique speeds up the search. Trajectories are represented as a bipartite graph, where nodes are trajectory IDs and edges indicate matching bounding boxes. Clustering is performed via breadth-first search (BFS)\cite{bundy1984breadth} to convert the bipartite graph into a forest.

\begin{figure}
    \hfill
    \begin{subfigure}[t]{0.3\linewidth}
        \centering
        \includegraphics[width=0.3\linewidth]{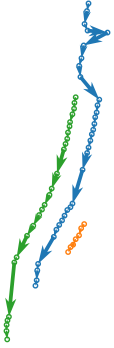}
        \caption{Forward tracking results}
    \end{subfigure}
    \hfill
    \begin{subfigure}[t]{0.3\linewidth}
        \centering
        \includegraphics[width=0.3\linewidth]{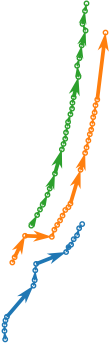}
        \caption{Backward tracking results}
    \end{subfigure}
    \hfill
    \begin{subfigure}[t]{0.3\linewidth}
        \centering
        \includegraphics[width=0.3\linewidth]{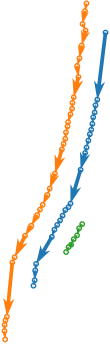}
        \caption{Bidirectional fusion results}
    \end{subfigure}

    \caption{Visualization of the bidirectional trajectory fusion. Circles represent detected boxes and arrows are directions.
    }
    \label{fig:track-fusion}
\end{figure}

\subsubsection{Bidirectional Trajectory Fusion}
This work identifies two common discrepancies in forward and backward tracking: (1) large velocity changes and (2) incorrect velocity initialization. The bidirectional fusion strategy has two main goals: (1) merge as many trajectory fragments as possible, and (2) prioritize fragments from longer trajectories. For clusters with no time contradiction, trajectories are directly merged; otherwise, they are split, selected, and merged at the fragment level. Common object links form guaranteed fragments, while others are candidates. Fragments are merged only if their time frames do not overlap. A greedy approach, based on the priority function \eqref{eq:bi-opt}, is used to select candidate fragments. A visual example is shown in Fig. \ref{fig:track-fusion}, with the full procedure in Algorithm \ref{algo:track-fusion}.


\subsubsection{Single-Trajectory Refinement}

After obtaining accurate object links, each single trajectory can be further refined via trajectory completion and smoothing. For trajectory completion, this work performs the linear interpolation on 3D positions and detection confidences to recover missed objects. To avoid adding false positives, this work only interpolates objects in a sliding window of size $\theta^\text{inter}$ and filters the results that are too close to existing objects according to NCD and the threshold $\gamma$. For trajectory smoothing, this work performs confidence-weighted size averaging for rigid objects and regresses 3D positions via the Gaussian process, whose radial basis function kernel $e^{-\frac{||x-x'||^2}{2 \sigma^2}}$ is based on the adaptive smoothness control function\cite{du2023strongsort}: $\sigma = \tau \log \frac{\tau^3}{|T_i|}$, where $|T_i|$ is the trajectory length and $\tau$ is a hyper-parameter.
\begin{algorithm}[!t]
    \scriptsize
    \caption{Bidirectional multi-trajectory fusion}
    \label{algo:track-fusion}
    \KwIn{\parbox[t]{\linewidth}{
    Tracking ID clusters $\boldsymbol{\mathcal{S}}$;\\
    Forward trajectories $\boldsymbol{\mathcal{T}}^\text{A}$ and object links $\boldsymbol{\mathcal{L}}^\text{A}$;\\
    Backward trajectories $\boldsymbol{\mathcal{T}}^\text{B}$ and object links $\boldsymbol{\mathcal{L}}^\text{B}$;\\
    Common boxes between trajectories $\boldsymbol{\mathcal{O}}$;
    \vspace{1mm}
    }}
    \KwOut{Fused trajectories $\boldsymbol{\mathcal{T}}^\text{C} $;}
    $\boldsymbol{\mathcal{T}}^\text{C} \gets \varnothing$\;
    \ForEach{$\boldsymbol{S}_i \in \boldsymbol{\mathcal{S}}$}{
        \eIf{$|\boldsymbol{S}_i| = 1$}{
            $\boldsymbol{\mathcal{L}}^{(\boldsymbol{S}_i)} \gets \left\{L_j \mid \text{id}(L_j) \in \boldsymbol{S}_i, L_j \in \boldsymbol{\mathcal{L}}^\text{A} \cup \boldsymbol{\mathcal{L}}^\text{B} \right \}$\;
        }{
            $\boldsymbol{\mathcal{O}}^{(\boldsymbol{S}_i)} \gets \left \{O_j \mid \text{id}(O_j) \in \boldsymbol{S}_i, O_j \in \boldsymbol{\mathcal{O}} \right \}$\;
            \eIf{$\boldsymbol{\mathcal{O}}^{(\boldsymbol{S}_i)}$ \upshape are in different frames}{
                $\boldsymbol{\mathcal{L}}^{(\boldsymbol{S}_i)} \gets $ Link $\boldsymbol{\mathcal{O}}^{(\boldsymbol{S}_i)}$ together\;
            }{
                $\boldsymbol{\mathcal{T}}^{\text{A}(\boldsymbol{S}_i)} \gets \left\{T_j \mid \text{id}(T_j) \in \boldsymbol{S}_i, T_j \in \boldsymbol{\mathcal{T}}^\text{A} \right\}$\;
                $\boldsymbol{\mathcal{T}}^{\text{B}(\boldsymbol{S}_i)} \gets \left\{T_j \mid \text{id}(T_j) \in \boldsymbol{S}_i, T_j \in \boldsymbol{\mathcal{T}}^\text{B} \right\}$\;
                $\boldsymbol{\mathcal{F}}^{\text{A}(\boldsymbol{S}_i)}, \boldsymbol{\mathcal{F}}^{\text{B}(\boldsymbol{S}_i)} \gets$ Split $\boldsymbol{\mathcal{T}}^{\text{A}(\boldsymbol{S}_i)}, \boldsymbol{\mathcal{T}}^{\text{B}(\boldsymbol{S}_i)}$ using $ \boldsymbol{\mathcal{L}}^{\text{A}(\boldsymbol{S}_i)} \cap \boldsymbol{\mathcal{L}}^{\text{B}(\boldsymbol{S}_i)}$\;
                $\boldsymbol{\mathcal{F}}^{(\boldsymbol{S}_i)} \gets \boldsymbol{\mathcal{F}}^{\text{A}(\boldsymbol{S}_i)} \cap \boldsymbol{\mathcal{F}}^{\text{B}(\boldsymbol{S}_i)}$\;
                $\boldsymbol{\mathcal{M}} \gets \boldsymbol{\mathcal{F}}^{\text{A}(\boldsymbol{S}_i)} \cup \boldsymbol{\mathcal{F}}^{\text{B}(\boldsymbol{S}_i)} - \boldsymbol{\mathcal{F}}^{(\boldsymbol{S}_i)}$\;
                $\boldsymbol{\mathcal{H}} = \left\{h(F_j) \mid F_j \in \boldsymbol{\mathcal{M}} \right \} $\;
                $\boldsymbol{\mathcal{M}} \gets $ Sort $\boldsymbol{\mathcal{M}}$ by $\boldsymbol{\mathcal{H}}$ in descending order\;
                \ForEach{$F_j \in \boldsymbol{\mathcal{M}}$}{
                    $\boldsymbol{\mathcal{N}} \gets \{F_k \mid \text{id}(F_j) = \text{id}(F_k), F_k \in \boldsymbol{\mathcal{F}}^{(\boldsymbol{S}_i)} \} \cup \{F_j \} $\;
                    \If{\upshape the objects of $\boldsymbol{\mathcal{N}}$ \upshape are in different frames}{
                        $\boldsymbol{\mathcal{F}}^{(\boldsymbol{S}_i)} \gets \boldsymbol{\mathcal{F}}^{(\boldsymbol{S}_i)} \cup \{F_j\}$
                    }
                }
            }
        }
        $\boldsymbol{\mathcal{T}}^\text{C} \gets \boldsymbol{\mathcal{T}}^\text{C} \cup \boldsymbol{\mathcal{F}}^{(\boldsymbol{S}_i)}$;
    }
    \Return $\boldsymbol{\mathcal{T}}^\text{C} $
\end{algorithm}
\section{Experiments}
\label{sec:experiments}

\begin{table*}
    \caption{Comparison with PC3T\cite{wu20213d} on the KITTI training set.}
    \label{tab:pc3t}
    \centering
    \begin{threeparttable}[c]
    \begin{tabularx}{\linewidth}{A | A A A A | A A A }
        \toprule
         & FP$\downarrow$ & FN$\downarrow$ & IDSW$\downarrow$ & \textbf{MOTA}$\uparrow$ & DetA$\uparrow$ & AssA$\uparrow$ & \textbf{HOTA}$\uparrow$ \\
        \midrule
        PC3T & 1352 & 1959 & 13 & 86.19 \% & 79.95 \% & 86.87 \% & 83.28 \% \\
        PC3T\tnote{+} & 18 & 1195 & 496 & 86.62 \% & 80.30 \% & 86.63 \% & 83.34 \% \\
        PC3T\tnote{++} & 14 & 1202 & 494 & 86.62 \% & 80.67 \% & 87.37 \% & 83.86 \% \\
        BiTrack (ours) & 1266 & 1655 & 13 & \textbf{87.81} \% & 81.93 \% & 87.38 \% & \textbf{84.54} \% \\
        \bottomrule
    \end{tabularx}
    \begin{tablenotes}
        \item \tnote{+} using our 2D-3D detection fusion module; \tnote{++} using our 2D-3D detection fusion and single-trajectory refinement modules.
    \end{tablenotes}
    \end{threeparttable}
    \vspace{-0.3cm} 
\end{table*}

\begin{table*}
    \caption{Comparison with state-of-the-art 3D OMOT methods on the KITTI test set.}
    \label{tab:test}
    \centering
    \begin{threeparttable}[c]
    \begin{tabularx}{\linewidth}{C{0.15\linewidth} | A A A A | A A A}
        \toprule
        & FP$\downarrow$ & FN$\downarrow$ & IDSW$\downarrow$ & \textbf{MOTA}$\uparrow$ & DetA$\uparrow$ & AssA$\uparrow$ & \textbf{HOTA}$\uparrow$ \\
        \midrule
        CasTrack\cite{CasA,wu20213d} & 1227 & 1533 & 24 & \textbf{91.91} \% & 78.58 \% & 84.22 \% & 81.00 \% \\
        VirConvTrack\cite{Wu_2023_CVPR,wu20213d} & 702 & 2648 & 8 & 90.24 \% & 78.14 \% & 86.39 \% & 81.87 \% \\
        BiTrack (ours) & 959 & 1892 & 21 & 91.65 \% & 80.13 \% & 86.07 \% & \textbf{82.69} \% \\
        \bottomrule
    \end{tabularx}
    \begin{tablenotes}
        \item Only cars are evaluated. Results are available at \url{https://www.cvlibs.net/datasets/kitti/eval_tracking.php}
    \end{tablenotes}
    \end{threeparttable}
    \vspace{-0.5cm} 
\end{table*}

\subsection{Dataset and Evaluation Metrics}
The KITTI tracking benchmark\cite{Geiger2012CVPR} was used as the evaluation platform. It includes 21 training sequences and 29 test sequences of front-view camera images and LiDAR point clouds. All sensor data are pre-calibrated and synchronized. The ground truth (GT) consists of 3D bounding boxes, categories, and tracking IDs. The experiments followed the official evaluation setting of the KITTI benchmark and the labels of cars in all training sequences were used. The judgement of true positive (TP), false positive (FP) and false negative (FN) for object detection is based on the rotated IoU in the 3D space, while the judgement for MOT is based on the 2D IoU on the image plane. The 3D IoU threshold for detection is 0.7 and the 2D IoU threshold for tracking is 0.5. The average precision (AP) metric with 40 recall positions is used for object detection.
MOTA\cite{MOTA} and HOTA\cite{Luiten2020ijCV} metrics are two main metrics for MOT, where MOTA penalizes FP, FN, and ID switches (IDSW) while HOTA relies on the detection accuracy (DetA) and the association accuracy (AssA).

\subsection{Implementation Details}
The 3D object detection method VirConv\cite{Wu_2023_CVPR} and the 2D instance segmentation method PointTrack\cite{pointtrack} were applied in the pre-processing to generate high-quality detection input. In the 2D-3D object fusion, this work set $\theta^\text{det} = 0.85$ and $\alpha = 0$. In the 3D trajectory generation, this work set $\beta = 0.5$, the initial state covariance $\boldsymbol{P}_0 = 10 \mathbf{I}$, the process covariance $\boldsymbol{Q} = 2 \mathbf{I}$, and the measurement covariance $\boldsymbol{R} = \mathbf{I}$, where $\mathbf{I}$ is the identity matrix. In the track management, this work set $\theta^\text{hit} = 6$ and $\theta^\text{miss} = \{5, 28\}$ for new trajectories and confirmed trajectories respectively. In the single-trajectory refinement, this work set $\theta^\text{inter} = 4$, $\gamma = 0.35$, and $\tau = 5.5$.  

\subsection{Comparative Results}
This work compared to a strong OMOT baseline PC3T\cite{wu20213d}, which also uses the tracking-by-detection paradigm based on the 3D Kalman filter. Both BiTrack and PC3T need no network training process for tracking, thus this work used all 21 training sequences for evaluation. To perform fair comparison, all experiments used the same 3D object detection source (VirConv\cite{Wu_2023_CVPR}). TABLE \ref{tab:pc3t} shows the superiority of BiTrack in terms of both MOTA (+1.62\%) and HOTA (+1.26\%). To further demonstrate the effectiveness of the proposed methods, this work re-implemented the baseline with the proposed 2D-3D detection fusion module and the single-trajectory refinement module based on the official code. The results show that the both two modules produce performance gains for PC3T but BiTrack is still better than the improved baseline. 

This work submitted the test-set results to the KITTI test server. TABLE \ref{tab:test} shows the MOT result comparison on the KITTI test set, where BiTrack outperforms all the other public MOT methods in terms of HOTA. The main superiority falls in the detection accuracy (DetA) due to the relatively fewer false alarms and missed objects. The association accuracy (AssA) is a bit lower than VirConvTrack\cite{Wu_2023_CVPR,wu20213d}, which may due to the larger number of detected objects.

\subsection{Ablation Studies}

For 2D-3D object fusion, this work compared four object representations: (1) 3D bounding boxes, (2) 2D bounding boxes, (3) 3D object points, and (4) 2D instance masks, focusing on detection precision and tracking accuracy. The 2D bounding box was derived from 2D instance masks, with the number of inside-box pixels used as the similarity metric for both “box-point” and “mask-box” fusion. All methods shared the same MOT module and detection confidence threshold. TABLE \ref{tab:2d-3d-fusion} shows that the proposed “mask-point” fusion outperforms other box-level methods in both detection precision and tracking accuracy for hard objects.

\begin{table}[!t]
    \caption{Evaluation of different object representations and registration cues for 2D-3D object fusion.}
    \label{tab:2d-3d-fusion}
    \centering
    \begin{threeparttable}[c]
    \begin{tabularx}{\linewidth}{A A | A A A A}
        \toprule
        2D Obj. & 3D Obj. & AP$\uparrow$ (mod.) & AP$\uparrow$ (hard) & MOTA$\uparrow$ & HOTA$\uparrow$\\
        \midrule
        $\times$ & Box & 96.01 \% & 95.69 \% & 86.27 \% & 83.03 \% \\
        Box & Box & 96.32 \% & 95.99 \% & 87.40 \% & 83.45 \% \\
        Mask & Box & \textbf{96.49} \% & 93.94 \% & 83.94 \% & 81.14 \% \\
        Box & Point & 96.32 \% & 96.00 \% & 87.31 \% & 83.40 \% \\
        Mask & Point & 96.33 \% & \textbf{96.01} \% & \textbf{87.41} \% & \textbf{83.46} \% \\
        \bottomrule
    \end{tabularx}
    \end{threeparttable}
\vspace{-0.5cm}
\end{table}

For the initial trajectory generation, the proposed improvements for the Kalman filter-based 3D MOT were evaluated. TABLE \ref{tab:3d-mot} shows that all the modifications have positive effects on tracking accuracy, especially for the NCD similarity metric and the track recovery. The double miss thresholds and the velocity re-initialization gives small improvements. 

\begin{table}[h]
    \caption{Evaluation of the different similarity metrics and track management mechanisms for 3D trajectory generation.}
    \label{tab:3d-mot}
    \centering
    \begin{threeparttable}[c]
    \begin{tabularx}{\linewidth}{A A A A | A A}
        \toprule
        $\boldsymbol{\mathcal{W}}^\text{T}$ & Track Recov. & $\theta^\text{miss}$ & Velocity Re-Init. & MOTA$\uparrow$ & HOTA$\uparrow$\\
        \midrule
        IoU & $\times$ & 28 & $\times$ & 86.67 \% & 82.87 \% \\
        CD & $\times$ & 28 & $\times$ & 86.22 \% & 82.53 \% \\
        NCD & $\times$ & 28 & $\times$ & 86.88 \% & 83.30 \% \\
        NCD & \Checkmark & 28 & $\times$ & 87.31 \% & 83.54 \% \\
        NCD & \Checkmark & \{5, 28\} & $\times$ & 87.32 \% & \textbf{83.55} \% \\
        NCD & \Checkmark & \{5, 28\} & \Checkmark & \textbf{87.37} \% & \textbf{83.55} \% \\
        \bottomrule
    \end{tabularx}
    \end{threeparttable}
\end{table}

The bidirectional multi-trajectory fusion module was evaluated based on the best unidirectional tracking results. TABLE \ref{tab:forward-backward-fusion} shows that the backward trajectories are slightly less accurate than the forward trajectories in average. However, the proposed method can improve the trajectories by selecting the better object links between them. 

\begin{table}[!t]
    \caption{Evaluation of the proposed bidirectional multi-trajectory fusion method.}
    \label{tab:forward-backward-fusion}
    \centering
    \begin{threeparttable}[c]
    \begin{tabularx}{\linewidth}{A | A A }
        \toprule
         & MOTA$\uparrow$ & HOTA$\uparrow$\\
        \midrule
        forward MOT & 87.37 \% & 83.55 \% \\
        backward MOT & 87.32 \% & 83.37  \% \\
        bidirectional fusion & \textbf{87.38} \% & \textbf{83.59} \% \\
        \bottomrule
    \end{tabularx}
    \end{threeparttable}
\end{table}

The single trajectory refinement module evaluated weighted size averaging, linear interpolation, and Gaussian process regression with bidirectional fusion results. TABLE \ref{tab:refinement} shows that all methods enhanced performance by generating more complete trajectories and improving object regression accuracy.

\begin{table}[!t]
    \caption{Evaluation of the proposed single-trajectory refinement method.}
    \label{tab:refinement}
    \centering
    \begin{threeparttable}[c]
    \begin{tabularx}{\linewidth}{A A A | A A }
        \toprule
        Weighted Size Avg. & Linear Interp. & Gaussian Process & MOTA$\uparrow$ & HOTA$\uparrow$ \\
        \midrule
        $\times$ & $\times$ & $\times$ & 87.38 \% & 83.59 \% \\
        \Checkmark & $\times$ & $\times$ & 87.46 \% & 84.23 \% \\
        \Checkmark & \Checkmark & $\times$ & 87.77 \% & 84.43 \% \\
        \Checkmark & \Checkmark & \Checkmark & \textbf{87.81} \% & \textbf{84.54} \% \\
        \bottomrule
    \end{tabularx}
    \end{threeparttable}
    \vspace{-0.5cm} 
\end{table}

\section{Conclusion}
\label{sec:conclusion}
This paper introduces BiTrack, an advanced OMOT framework that integrates 2D-3D detection fusion, bidirectional tracking, and trajectory re-optimization. By leveraging point-level correspondences, an object motion similarity metric, enhanced hit-miss management, fragment fusion techniques, and physical models, BiTrack ensures the accurate generation of offline trajectories. Demonstrating state-of-the-art performance on the KITTI leaderboard, BiTrack, with its modular architecture and fully automated pipeline, is highly adaptable for use in 3D object annotation platforms and various other offline tasks.
\nocite{*}
\bibliographystyle{IEEEtran}
\bibliography{IEEEabrv,refs}

\begin{thebibliography}{10}
\providecommand{\url}[1]{#1}
\csname url@samestyle\endcsname
\providecommand{\newblock}{\relax}
\providecommand{\bibinfo}[2]{#2}
\providecommand{\BIBentrySTDinterwordspacing}{\spaceskip=0pt\relax}
\providecommand{\BIBentryALTinterwordstretchfactor}{4}
\providecommand{\BIBentryALTinterwordspacing}{\spaceskip=\fontdimen2\font plus
\BIBentryALTinterwordstretchfactor\fontdimen3\font minus \fontdimen4\font\relax}
\providecommand{\BIBforeignlanguage}[2]{{%
\expandafter\ifx\csname l@#1\endcsname\relax
\typeout{** WARNING: IEEEtran.bst: No hyphenation pattern has been}%
\typeout{** loaded for the language `#1'. Using the pattern for}%
\typeout{** the default language instead.}%
\else
\language=\csname l@#1\endcsname
\fi
#2}}
\providecommand{\BIBdecl}{\relax}
\BIBdecl

\bibitem{bewley2016simple}
A.~Bewley, Z.~Ge, L.~Ott, F.~Ramos, and B.~Upcroft, ``Simple online and realtime tracking,'' in \emph{2016 IEEE International Conference on Image Processing (ICIP)}, 2016, pp. 3464--3468.

\bibitem{Weng2020_AB3DMOT}
X.~Weng, J.~Wang, D.~Held, and K.~Kitani, ``{3D Multi-Object Tracking: A Baseline and New Evaluation Metrics},'' \emph{IROS}, 2020.

\bibitem{wu20213d}
H.~Wu, W.~Han, C.~Wen, X.~Li, and C.~Wang, ``3d multi-object tracking in point clouds based on prediction confidence-guided data association,'' \emph{IEEE Transactions on Intelligent Transportation Systems}, vol.~23, no.~6, pp. 5668--5677, 2022.

\bibitem{du2023strongsort}
Y.~Du, Z.~Zhao, Y.~Song, Y.~Zhao, F.~Su, T.~Gong, and H.~Meng, ``Strongsort: Make deepsort great again,'' \emph{IEEE Transactions on Multimedia}, pp. 1--14, 2023.

\bibitem{huang2021joint}
K.~Huang and Q.~Hao, ``Joint multi-object detection and tracking with camera-lidar fusion for autonomous driving,'' in \emph{2021 IEEE/RSJ International Conference on Intelligent Robots and Systems (IROS)}, 2021, pp. 6983--6989.

\bibitem{networkflows}
L.~Zhang, Y.~Li, and R.~Nevatia, ``Global data association for multi-object tracking using network flows,'' in \emph{2008 IEEE Conference on Computer Vision and Pattern Recognition}, 2008, pp. 1--8.

\bibitem{mussp}
C.~Wang, Y.~Wang, Y.~Wang, C.-T. Wu, and G.~Yu, ``mussp: Efficient min-cost flow algorithm for multi-object tracking,'' in \emph{Advances in Neural Information Processing Systems}, H.~Wallach, H.~Larochelle, A.~Beygelzimer, F.~d\textquotesingle Alch\'{e}-Buc, E.~Fox, and R.~Garnett, Eds., vol.~32.\hskip 1em plus 0.5em minus 0.4em\relax Curran Associates, Inc., 2019.

\bibitem{AHC_ETE}
K.~Zhao, T.~Imaseki, H.~Mouri, E.~Suzuki, and T.~Matsukawa, ``From certain to uncertain: Toward optimal solution for offline multiple object tracking,'' in \emph{2020 25th International Conference on Pattern Recognition (ICPR)}, 2021, pp. 2506--2513.

\bibitem{ReMOT}
F.~Yang, X.~Chang, S.~Sakti, Y.~Wu, and S.~Nakamura, ``Remot: A model-agnostic refinement for multiple object tracking,'' \emph{Image and Vision Computing}, vol. 106, p. 104091, 2021.

\bibitem{TMOH}
D.~Stadler and J.~Beyerer, ``Improving multiple pedestrian tracking by track management and occlusion handling,'' in \emph{2021 IEEE/CVF Conference on Computer Vision and Pattern Recognition (CVPR)}, 2021, pp. 10\,953--10\,962.

\bibitem{Qi_2021_CVPR}
C.~R. Qi, Y.~Zhou, M.~Najibi, P.~Sun, K.~Vo, B.~Deng, and D.~Anguelov, ``Offboard 3d object detection from point cloud sequences,'' in \emph{Proceedings of the IEEE/CVF Conference on Computer Vision and Pattern Recognition (CVPR)}, June 2021, pp. 6134--6144.

\bibitem{shenoi2020jrmot}
A.~{Shenoi}, M.~{Patel}, J.~{Gwak}, P.~{Goebel}, A.~{Sadeghian}, H.~{Rezatofighi}, R.~{Mart\'in-Mart\'in}, and S.~{Savarese}, ``Jrmot: A real-time 3d multi-object tracker and a new large-scale dataset,'' in \emph{2020 IEEE/RSJ International Conference on Intelligent Robots and Systems (IROS)}, 2020, pp. 10\,335--10\,342.

\bibitem{pang2020clocs}
S.~Pang, D.~Morris, and H.~Radha, ``Clocs: Camera-lidar object candidates fusion for 3d object detection,'' in \emph{2020 IEEE/RSJ International Conference on Intelligent Robots and Systems (IROS)}, 2020, pp. 10\,386--10\,393.

\bibitem{Qi_2018_CVPR}
C.~R. Qi, W.~Liu, C.~Wu, H.~Su, and L.~J. Guibas, ``Frustum pointnets for 3d object detection from rgb-d data,'' in \emph{Proceedings of the IEEE Conference on Computer Vision and Pattern Recognition (CVPR)}, June 2018.

\bibitem{kuhn1955hungarian}
H.~W. Kuhn, ``The hungarian method for the assignment problem,'' \emph{Naval research logistics quarterly}, vol.~2, no. 1-2, pp. 83--97, 1955.

\bibitem{Wu_2023_CVPR}
H.~Wu, C.~Wen, S.~Shi, X.~Li, and C.~Wang, ``Virtual sparse convolution for multimodal 3d object detection,'' in \emph{Proceedings of the IEEE/CVF Conference on Computer Vision and Pattern Recognition (CVPR)}, June 2023, pp. 21\,653--21\,662.

\bibitem{Geiger2012CVPR}
A.~Geiger, P.~Lenz, and R.~Urtasun, ``Are we ready for autonomous driving? the kitti vision benchmark suite,'' in \emph{2012 IEEE Conference on Computer Vision and Pattern Recognition}, 2012, pp. 3354--3361.

\bibitem{weng20203d}
X.~Weng, J.~Wang, D.~Held, and K.~Kitani, ``3d multi-object tracking: A baseline and new evaluation metrics,'' in \emph{2020 IEEE/RSJ International Conference on Intelligent Robots and Systems (IROS)}, 2020, pp. 10\,359--10\,366.

\bibitem{bundy1984breadth}
A.~Bundy and L.~Wallen, ``Breadth-first search,'' \emph{Catalogue of artificial intelligence tools}, pp. 13--13, 1984.

\bibitem{CasA}
H.~Wu, J.~Deng, C.~Wen, X.~Li, C.~Wang, and J.~Li, ``Casa: A cascade attention network for 3-d object detection from lidar point clouds,'' \emph{IEEE Transactions on Geoscience and Remote Sensing}, vol.~60, pp. 1--11, 2022.

\bibitem{MOTA}
K.~Bernardin and R.~Stiefelhagen, ``Evaluating multiple object tracking performance: The clear mot metrics,'' \emph{EURASIP Journal on Image and Video Processing}, vol. 2008, no.~1, p. 246309, 2008.

\bibitem{Luiten2020ijCV}
J.~Luiten, A.~Osep, P.~Dendorfer, P.~Torr, A.~Geiger, L.~Leal-Taixe, and B.~Leibe, ``Hota: A higher order metric for evaluating multi-object tracking,'' \emph{International Journal of Computer Vision}, vol. 129, no.~2, pp. 548--578, 2021.

\bibitem{pointtrack}
Z.~Xu, W.~Yang, W.~Zhang, X.~Tan, H.~Huang, and L.~Huang, ``Segment as points for efficient and effective online multi-object tracking and segmentation,'' \emph{IEEE Transactions on Pattern Analysis and Machine Intelligence}, vol.~44, no.~10, pp. 6424--6437, 2022.

\bibitem{wojke2017simple}
N.~Wojke, A.~Bewley, and D.~Paulus, ``Simple online and realtime tracking with a deep association metric,'' in \emph{2017 IEEE International Conference on Image Processing (ICIP)}, 2017, pp. 3645--3649.

\bibitem{deng2021voxel}
J.~Deng, S.~Shi, P.~Li, W.~Zhou, Y.~Zhang, and H.~Li, ``Voxel r-cnn: Towards high performance voxel-based 3d object detection,'' \emph{Proceedings of the AAAI Conference on Artificial Intelligence}, vol.~35, no.~2, pp. 1201--1209, May 2021.

\bibitem{spatialembedding}
D.~Neven, B.~D. Brabandere, M.~Proesmans, and L.~Van~Gool, ``Instance segmentation by jointly optimizing spatial embeddings and clustering bandwidth,'' in \emph{2019 IEEE/CVF Conference on Computer Vision and Pattern Recognition (CVPR)}, 2019, pp. 8829--8837.

\bibitem{openpcdet2020}
O.~D. Team, ``Openpcdet: An open-source toolbox for 3d object detection from point clouds,'' \url{https://github.com/open-mmlab/OpenPCDet}, 2020.

\bibitem{ijcai2021p161}
H.~Wu, Q.~Li, C.~Wen, X.~Li, X.~Fan, and C.~Wang, ``Tracklet proposal network for multi-object tracking on point clouds,'' in \emph{Proceedings of the Thirtieth International Joint Conference on Artificial Intelligence, {IJCAI-21}}, Z.-H. Zhou, Ed.\hskip 1em plus 0.5em minus 0.4em\relax International Joint Conferences on Artificial Intelligence Organization, 8 2021, pp. 1165--1171, main Track.

\bibitem{pmlr-v162-tokmakov22a}
P.~Tokmakov, A.~Jabri, J.~Li, and A.~Gaidon, ``Object permanence emerges in a random walk along memory,'' in \emph{Proceedings of the 39th International Conference on Machine Learning}, ser. Proceedings of Machine Learning Research, K.~Chaudhuri, S.~Jegelka, L.~Song, C.~Szepesvari, G.~Niu, and S.~Sabato, Eds., vol. 162.\hskip 1em plus 0.5em minus 0.4em\relax PMLR, 17--23 Jul 2022, pp. 21\,506--21\,519.

\bibitem{wu2022casa}
H.~Wu, J.~Deng, C.~Wen, X.~Li, C.~Wang, and J.~Li, ``Casa: A cascade attention network for 3-d object detection from lidar point clouds,'' \emph{IEEE Transactions on Geoscience and Remote Sensing}, vol.~60, pp. 1--11, 2022.

\end{thebibliography}


 





\end{document}